\documentclass[sigconf,nonacm]{acmart}
\makeatletter
\providecommand{\@LN}[2]{}
\providecommand{\@LN@col}[1]{}
\makeatother
\AtBeginDocument{%
  }
\usepackage{graphicx}
\usepackage{booktabs}
\usepackage{listings}
\usepackage{tikz}
\usetikzlibrary{positioning,arrows.meta}
\lstdefinestyle{prog}{
  basicstyle=\ttfamily\scriptsize,
  breaklines=true,
  columns=fullflexible,
  keepspaces=true,
  frame=single,
  framesep=3pt,
  xleftmargin=3pt,
  showstringspaces=false,
  aboveskip=5pt,
  belowskip=3pt
}
\begin{document}
\title[LiveSearch: A Live Financial Search Benchmark]{LiveSearch: Questions as Executable Verifier Code for a Live Benchmark of Professional-Source Financial Search Agents}
%
% Author names feed the PDF metadata; the visible author block is typeset
% below with numbered affiliations instead of acmart's per-author grid.
\author{Sinuo Wang}
\author{Wentao Zhao}
\author{Jiacheng Lu}
\author{Rui Sun}
\author{Beidi Luan}
\author{Qiming Shi}
\author{Jing Li}
\author{Daxin Jiang}
\author{Zuo Bai}
\renewcommand{\shortauthors}{Wang et al.}
\newcommand{\au}[2]{\mbox{#1\textsuperscript{#2}}}
\newcommand{\livesearchauthors}{%
  {\LARGE\bfseries
    \au{Sinuo Wang}{1,3}, \au{Wentao Zhao}{4}, \au{Jiacheng Lu}{1,5},
    \au{Rui Sun}{1}, \au{Beidi Luan}{1},\\
    \au{Qiming Shi}{1,6}, \au{Jing Li}{1}, \au{Daxin Jiang}{1},
    \au{Zuo Bai}{1,2,\dag}\par}
  \bigskip
  {\Large
    \textsuperscript{1}StepFun\qquad
    \textsuperscript{2}FinStep\qquad
    \textsuperscript{3}Adelaide University\\
    \textsuperscript{4}Tsinghua University\qquad
    \textsuperscript{5}Shanghai Jiao Tong University\qquad
    \textsuperscript{6}Zhejiang University\par}
  \medskip
  {\large\textsuperscript{\dag}Corresponding author:
    \href{mailto:baizuo@stepfun.com}{baizuo@stepfun.com}\par}}
\makeatletter
\def\@mkauthors{\begingroup
  \hsize=\textwidth
  \global\setbox\mktitle@bx=\vbox{\noindent
    \unvbox\mktitle@bx\par\medskip
    {\centering\livesearchauthors\par}\bigskip}%
  \endgroup}
\makeatother
\begin{abstract}
In finance, the answer that matters is usually the current one: prices, rates,
and disclosures hold only as of a given day, so retrieving up-to-date facts is
a core capability of a financial agent. Yet most financial benchmarks overlook
timeliness. They fix each answer at annotation time, so the gold goes stale as
markets move and, once leaked into training data, rewards recall rather than
search and analysis. We introduce \emph{LiveSearch}, a live benchmark that
stores each question as \emph{executable verifier code}: a typed program of
calls to a professional financial data service, followed by a declarative
extractor. The question is frozen, while its gold answer is determined by
executing the program against professional financial data at evaluation
time, and a daily job re-executes it with no human or language model in the
loop. A memorized answer is thus likely
outdated, and a model must reach the current one through tools and analysis.
Construction is equally automatic: a language model only \emph{verbalizes}
programs enumerated from real tool responses and never chooses an answer.
Under general web search, six contemporary tool-calling models reach
$70.9\%$--$76.7\%$ answer accuracy. Over three consecutive days, $70\%$ of
served questions change their gold value, and even under tolerance-aware
matching, a literally reused earlier snapshot fails on a majority of
questions within a single day. Per-question tracking uncovers dynamic shifts in correctness masked by aggregate accuracy.
\end{abstract}
\maketitle

\section{Introduction}
\label{sec:intro}

Large language models are now routinely deployed as \emph{search agents}:
given a natural-language question, the model plans a sequence of tool calls,
reads the returned evidence, and composes an answer
\cite{react,toolllm,webarena, wangevolving}. Finance is a natural and high-value proving
ground for such agents (questions are precise, answers are checkable, and the
underlying data is abundant and structured), and the field has moved from
domain-specific models and static task suites
\cite{bloomberggpt,pixiu,finben,finqa,convfinqa,financebench} to agentic
evaluation in which the model must retrieve its own evidence through search and
tools \cite{fingaia,finsearchcomp,financeagentbench,sun2025finresearchbenchlogictreebased,luan2026finresearchbenchiideepresearch,fintoolbench,frontierfinance,fintrace}.
Finance is also, more than most domains, \emph{intrinsically time-sensitive}:
the answer an analyst actually needs (a latest price, an interest-rate spread, a
newly filed fundamental, the current top holder) changes by the day or the
minute, and a value that was correct at annotation time is often wrong a day
later. What matters is therefore not only whether an agent can retrieve a fact,
but whether it can retrieve the \emph{current} one; the natural unit of ground
truth in finance is a value that is re-established over time rather than
written down once. Current benchmarks give up exactly this property. Despite
differing in language, size, tool inventory, and task taxonomy, essentially all
of these agentic benchmarks
\cite{fingaia,finsearchcomp,mcpatlas,livemcp,financeagentbench,fintoolbench,frontierfinance}
share three weaknesses that turn out to be one weakness seen from three angles.

\begin{enumerate}
\item \textbf{Curation is costly.}
Constructing financial agent benchmarks requires hundreds of
expert-hours~\cite{fingaia,finsearchcomp,financeagentbench,frontierfinance},
and without reusable answer-generation programs, every refresh repeats
annotation and validation.

\item \textbf{Static answers become stale.}
For time-sensitive queries, frozen reference answers silently go out of
date and, once leaked, reward memorization rather than fresh
retrieval~\cite{livebench,evobrowsecomp,zhou2025livesearchbench}.
FinSearchComp mitigates this with scripted updates, but only for its
time-sensitive tier~\cite{finsearchcomp}.

\item \textbf{Evidence departs from professional practice.}
Professional financial questions are formulated around the data fields
and metric definitions of specialized databases. Benchmarks built
primarily from web-accessible content or free-tier
tools~\cite{financeagentbench,frontierfinance,fintoolbench,livemcp}
tend to underrepresent these information needs.
\end{enumerate}

These three problems are coupled because they all stem from the same design
choice: \emph{storing a question as a (text, answer) pair}. If the answer is a
literal string, then someone must author it (manual curation), it cannot change
(static decay), and its provenance is whatever the author happened to consult
(evidence gap).

These challenges call for a unified mechanism that connects question
construction, reference-answer generation, and ongoing updates.
LiveSearch represents each question as a triple
$(\text{text}, \pi, \tau)$, where $\text{text}$ is a fixed natural-language
prompt, $\pi$ is an executable program over a professional financial data
service, and $\tau$ specifies how a candidate answer is matched to the
program output. Each program comprises tool calls followed by a
declarative extractor that selects and formats a single value.
During construction, the pipeline enumerates candidate programs from
observed tool responses and uses a language model to verbalize them;
reference values are determined by execution.
The same programs are then re-executed daily to produce dated gold
snapshots without further language-model involvement.
This design derives questions from professional data fields and metric
definitions, makes reference answers traceable to explicit queries and
extraction rules, and updates time-sensitive answers while keeping
the question text fixed.

For web search agents, the challenge is not just finding a relevant page:
online evidence can mix stale and current values, reporting periods, and
metric definitions. LiveSearch evaluates ordinary web search and page
retrieval against reference answers refreshed from professional data,
testing whether agents can recover the current, precisely specified fact
from a changing Web rather than recall a once-correct answer.

\paragraph{Contributions}

\begin{enumerate}
\item \textbf{A live benchmark for financial search agents.}
Each question is paired with an executable reference program over
professional financial data tools and a fixed answer-matching rule.
A snapshot-and-freeze protocol binds each question to a timestamp and
a frozen gold value. We report six models' scores under web search and
measure answer accuracy and temporal changes in gold answers
(Sections~\ref{sec:design} and~\ref{sec:results}).

\item \textbf{An automated construction and refresh pipeline.}
The pipeline enumerates candidate programs from observed tool responses
and uses a language model to screen candidates and verbalize selected
programs, while reference answers are computed by execution.
Each question's program is then fixed, and subsequent pool selection
and daily refresh proceed without further language-model involvement
(Sections~\ref{sec:answerfirst} and~\ref{sec:daily}).

\item \textbf{An eight-category financial asset taxonomy.}
The taxonomy includes two cross-cutting axes and entity-sampling
sources for each subcategory, defining the benchmark's scope and
making explicit which financial domains the underlying data service
covers (Section~\ref{sec:taxonomy}).
\end{enumerate}

\section{Related Work}
\label{sec:related}

\paragraph{Finance agent benchmarks.}
Domain-specific models and static evaluation suites
\cite{bloomberggpt,pixiu,finben} and report-grounded QA resources such as
FinQA, ConvFinQA, and FinanceBench~\cite{finqa,convfinqa,financebench} made
financial competence a measurable target, but they evaluate reasoning over
frozen text and tables rather than interaction with live tools. The Finance
Agent Benchmark~\cite{financeagentbench} (537 expert-authored questions over US
equities and SEC filings) and FinGAIA~\cite{fingaia} (407 Chinese expert tasks
in a GAIA style) are fully manual and static; the former defends against
contamination with a private test split and post-cutoff documents rather than
with freshness. FinSearchComp~\cite{finsearchcomp} is the closest work: $635$
bilingual expert questions in three tiers, with a time-sensitive tier whose
\emph{reference values} are regenerated by fixed scripts after market close.
Its question pool is static and its complex tier is deliberately historical, so
time sensitivity and professional grounding are never combined.

Concurrent benchmarks push finance agents toward richer tool use and richer
scoring: FinToolBench pairs hundreds of runnable free-tier tools with
tool-required questions and audits call-level compliance, FrontierFinance
scores long-form analyst queries against thousands of expert-written,
source-attributed rubrics, and FinTrace grades whole tool-calling trajectories
rather than final answers \cite{fintoolbench,frontierfinance,fintrace}. These
works improve \emph{how} an agent's behaviour is measured and are complementary
to ours. Their item is still a (text, answer-or-rubric) pair written once by
humans or by a one-off synthesis pass, so the pool ages as the market moves,
refreshing it means re-annotating it, and the evidence is free-tier or
web-tier. LiveSearch stores the item as executable code over professional
data tools, so the gold answer is re-derived daily by execution, contamination
decay is measurable per question, and no human or language model decides a
gold answer in either curation or refresh. It thus occupies the quadrant
that is both time-sensitive \emph{and} professional-source-grounded.
Lu et al.~\cite{lu2026analystgroundtruthgrading} are complementary in another
respect: they benchmark agents on building discounted-cash-flow (DCF) models
and grade the built models against observed professional practice rather than
a single analyst's model, whereas each LiveSearch question does resolve to one
authoritative value.

\paragraph{Dynamic evaluation.}
LiveBench~\cite{livebench} limits contamination by periodically adding
questions drawn from recent sources and scoring them against objective ground
truth. EvoBrowseComp~\cite{evobrowsecomp} identifies the analogous leakage
problem for search agents, whose static questions can enter training data or be
solved from parametric memory, and addresses it by synthesizing fresh English
and Chinese questions through live-web traversal and periodically regenerating
the pool. LiveSearchBench~\cite{zhou2025livesearchbench} automates
construction from knowledge updates: it computes deltas between successive
Wikidata snapshots, synthesizes questions at three reasoning levels from the
changed triples, and validates each question's unique answer with a SPARQL
query. That query is the closest analogue to our gold program, but it
certifies a question minted from a recent update; keeping pace with the world
means generating new questions from the next delta. In all three, the unit of
refresh is the question set. LiveSearch
instead freezes the question text and refreshes the answer by re-executing its
program. Finance motivates this choice: ``the latest disclosed return on equity
of company X'' stays well-posed indefinitely while its answer changes every
reporting period. Keeping the question also makes a single question's answer
drift, and the decay of a memorized answer, observable over time.

\paragraph{Automated question synthesis for web agents.}
Outside finance, several works synthesize questions for search agents:
WebDancer's CRAWLQA and E2HQA, WebSailor's rare-entity knowledge-graph random
walks with obfuscation, and WebWalker's website-traversal
QA~\cite{webdancer,websailor,webwalker}. None is finance-specific and none
refreshes its answers, but together they show that high-quality agent questions
can be synthesized rather than hand-written. Program-to-question generation and
schema-driven synthetic QA have also been studied in semantic
parsing~\cite{qgsql,autoqa}. LiveSearch applies this idea to a setting where
the synthesized artifact is an \emph{executable program} whose answer is
verifiable against live professional data.

\paragraph{Judging.}
A claim-level LLM judge with conjunctive scoring has been adopted independently
by the Finance Agent Benchmark~\cite{financeagentbench},
LiveMCPBench~\cite{livemcp}, and MCP-Atlas~\cite{mcpatlas}. Because each of our
questions resolves to one authoritative value, our metric is the single-claim
case: a binary judgment against the gold value, with numeric-tolerance matching
following FinSearchComp. Our harness fixes the judge within each run and draws
it from a model family absent from the leaderboard, since LLM judges are known
to favour their own family and to show position and verbosity
biases~\cite{llmjudge}.

\section{Method}
\label{sec:method}

\begin{figure*}[t]
\centering
\begin{tikzpicture}[
  font=\footnotesize,
  node distance=3mm and 4.5mm,
  stepbox/.style={draw=black!55, rounded corners=2pt, align=center,
    text width=3.2cm, minimum height=1.15cm, inner sep=2.5pt},
  det/.style={stepbox, fill=blue!6},
  llm/.style={stepbox, fill=orange!18},
  arr/.style={-{Latex[length=1.8mm]}, semithick, black!70},
  stage/.style={font=\footnotesize\bfseries, anchor=south west, inner sep=1pt}]
\node[det] (probe) {\textbf{Probe \& coverage}\\entity $\times$ tool calls,\\
  web visibility 0--10};
\node[det, right=of probe] (enum) {\textbf{Enumerate programs}\\every extractor
  that executes on an observed response};
\node[llm, right=of enum] (review) {\textbf{Filter \& review}\\mechanical rules,
  then LLM keep/reject per signature};
\node[llm, right=of review] (verb) {\textbf{Verbalize}\\LLM writes the text;
  gold $=$ executed extractor};
\node[det, below=12mm of probe] (select) {\textbf{Select pool}\\drop signatures
  that every model solves};
\node[det, right=of select] (mint) {\textbf{Re-instantiate}\\fresh entities per
  level, template substitution};
\node[det, right=of mint] (refresh) {\textbf{Daily refresh}\\re-date window,
  re-execute $\pi$};
\node[det, right=of refresh] (snap) {\textbf{Dated gold snapshot}\\
  \texttt{gold\_answers\_<date>}\\scores web-search agents};
\node[stage] at ([yshift=1mm]probe.north west)
  {Stage 1: answer-first construction (once)};
\node[stage] at ([yshift=1mm]select.north west)
  {Stage 2: no language model};
\draw[arr] (probe) -- (enum);
\draw[arr] (enum) -- (review);
\draw[arr] (review) -- (verb);
\draw[arr] (verb.south) -- ++(0,-4mm) -| ([xshift=-3mm]select.west)
  -- (select.west);
\draw[arr] (select) -- (mint);
\draw[arr] (mint) -- (refresh);
\draw[arr] (refresh) -- (snap);
\end{tikzpicture}
\caption{Overview of LiveSearch curation. Orange steps call a language model;
blue steps are deterministic. The model only verbalizes a program whose
extractor is already fixed, and the gold answer is whatever the program returns
when executed. Stage~2 carries no language model: pool selection reuses
templates, and the daily refresh re-dates, re-executes, and writes a dated gold
snapshot. All steps share one execution layer and one extractor engine.}
\Description{Two-row pipeline. Stage 1: probe and coverage, enumerate
programs, filter and review, verbalize. Stage 2: select pool, re-instantiate,
daily refresh, dated gold snapshot.}
\label{fig:method}
\end{figure*}
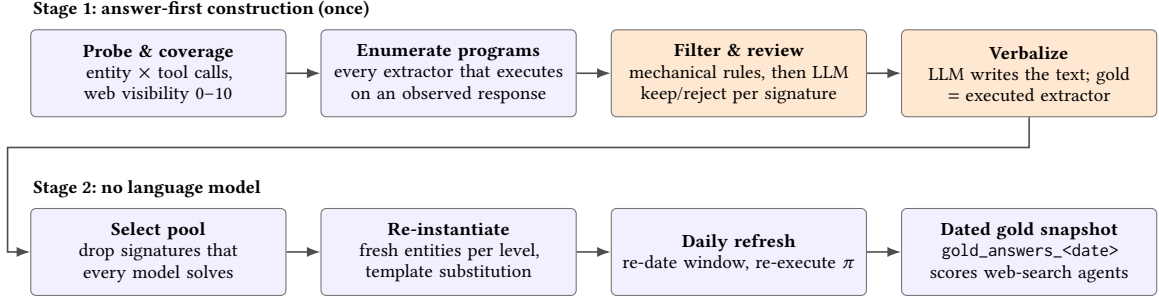

LiveSearch stores each question as a triple $(\text{text},\pi,\tau)$
and organizes construction and maintenance into two stages.
Section~\ref{sec:design} defines the question representation,
Section~\ref{sec:taxonomy} describes the underlying data service and
benchmark scope, and Sections~\ref{sec:answerfirst}
and~\ref{sec:daily} detail the two stages
(Figure~\ref{fig:method}).
Stage~1 builds an initial question inventory by querying financial
data tools, enumerating candidate programs from their responses,
and using a language model to screen candidates and generate
question text. Reference answers are computed by executing the programs.
Stage~2 selects the evaluation pool from this inventory and then
refreshes its reference answers daily, without further language-model
involvement. Both stages use the same tool execution layer and
extractor engine, so reference answers are generated and refreshed
through a shared implementation.

\subsection{Question Representation}
\label{sec:design}

\paragraph{Gold program.}
A question record stores the frozen question text, the executable \emph{gold
program}~$\pi$ whose execution defines the answer, the matching rule~$\tau$,
and metadata (Figure~\ref{fig:record}). The program is a short typed DAG: one
or more tool-call steps followed by a declarative \emph{extractor} that turns
the raw response into one formatted value. The extractor language has no
branches or expressions, so replaying a program is deterministic and every
program can be checked mechanically. An extractor names a path into the
response, a selection rule, a field, an optional aggregate, and a format (type,
decimals, scale, unit). The selection rule's \texttt{select.mode} $\in$
\{\texttt{latest}, \texttt{snapshot}, \texttt{match}, \texttt{minute},
\texttt{all}\} encodes the \emph{time anchor}: \texttt{latest} over a
\texttt{trade\_date} field means ``as of the most recent trading day,'' so the
selected row, and with it the answer, rolls forward as new data arrives;
\texttt{match} pins one row of a cross-sectional table by an identity key;
\texttt{snapshot} reads a single-record response. Each record also carries a
\texttt{refresh\_class} $\in$ \{daily, weekly, monthly, quarterly, annual,
static\}, the cadence at which its answer is expected to change.

\begin{figure}[t]
\begin{lstlisting}[style=prog]
{"qid": "a36fb1b0083",
 "question": "Brii Biosciences-B (02137.hk): back-adjusted
              price change on the latest trading day, in HKD?",
 "gold_program": [
  {"step": 1, "tool": "get_hk_stock_kline",
   "args": {"keyword": "02137.hk", "end_date": "2026-08-05",
            "reinstatement_type": 3}},
  {"step": 2, "extract": {
   "path": "data",
   "select": {"mode": "latest", "date_field": "trade_date",
              "where": {"stock_code": "02137.hk"}},
   "field": "price_change", "agg": null,
   "format": {"type": "number", "decimals": 6,
              "scale": 1, "unit": "HKD"}}}],
 "match_mode": "approx", "time_anchor": "latest trading day",
 "refresh_class": "daily"}
\end{lstlisting}
\caption{A question record (text glossed from Chinese). Each daily refresh
rewrites \texttt{end\_date} to the benchmark date and re-executes the program;
on 2026-09-16 it returned $-0.015$ HKD.}
\Description{JSON question record with qid, question text, a two-step gold
program (a K-line tool call and an extractor), matching mode, time anchor, and
refresh class.}
\label{fig:record}
\end{figure}

\paragraph{Matching rule.}
The matching rule~$\tau$ is one of \texttt{exact}, \texttt{approx}, or
\texttt{semantic}, and states how an answer is compared with the gold value.
\texttt{exact} rounds the answer to the gold value's precision;
\texttt{approx} accepts agreement at the coarser of the two precisions (at
least two decimals), an unrounded form of the gold value, or a relative error
within a fixed tolerance; and \texttt{semantic} accepts an equivalent meaning.
Both numeric
rules normalize units first, and a sign or direction mismatch fails under every
rule. The rule is fixed at construction: the language model proposes one, and
the pipeline clamps it mechanically, allowing \texttt{exact} only when nothing
marks the value as derived (a percentage unit, a tool display coarser than the
rendering, or a lossy rendering).

\paragraph{Time anchors, not absolute dates.}
An automated gate rejects any question whose text contains an absolute date,
including Chinese date and quarter forms. Questions refer to time only
relatively (``the most recent trading day,'' ``the latest disclosed reporting
period''), and the concrete window lives only in the program's arguments, which
the daily refresh rewrites. The same frozen sentence therefore has a new
correct answer every day the underlying series moves.

\paragraph{Snapshot-and-freeze.}
Because gold answers change by design, the unit of comparison is not a file but
a tuple: (question, timestamp, frozen gold value). Each daily refresh writes a
dated, immutable snapshot \texttt{gold\_answers\_<date>.jsonl} that records, per
question, the formatted answer, the raw value, the observed period, and the
arguments actually sent, so every gold value can be traced to the call that
produced it. An evaluation run cites the snapshot date it scored against.
Question identifiers are content hashes of the tool, entity, non-date
arguments, and extractor rather than positions, so a question keeps its
identity across regenerations. A same-date leaderboard comparison requires
identical qid sets and gold dates; a longitudinal audit instead fixes the qids
and explicitly tracks each date's own frozen gold values. The evaluator is
strict: the snapshot must
cover every served qid, contain no unknown qids, and carry no empty or
error-flagged answers, or the run fails rather than falling back to a stale
value.

\subsection{Data Source and Scope}
\label{sec:taxonomy}

\paragraph{Data source.}
Gold programs run on a production financial data service that exposes $96$
tools in $9$ services through the Model Context Protocol~\cite{mcp}, covering
company fundamentals and disclosures, quotes and microstructure, macro and rate
series, sector constituents, funds, indices, margin trading, utilities, and
news search. Tools take typed JSON arguments and return structured, dated
records. Three kinds of tool never define a gold program: search tools, so that
no gold value is sourced from generic search; intraday tools, whose
minute-level values would give two refreshes on the same day different answers;
and utilities such as the system clock.

\paragraph{Asset taxonomy.}
We organize the financial universe into a two-level taxonomy with eight
level-1 asset categories (Cash \& Monetary; Credit \& Financing; Equity \&
Ownership; Physical \& Alternative; Risk Transfer \& Derivatives; Composite \&
Structured Products; Foreign Exchange; and Indices, Benchmarks \& Reference
Rates), each split into level-2 subcategories, plus two cross-cutting axes.
Every subcategory names an \texttt{entity\_sampling\_source}, the registry or
feed from which real instruments can be drawn (e.g.\ SEC EDGAR CIK lists and
exchange listings for listed equity, CNINFO and ChinaBond for Chinese bonds,
AMAC for funds). The two axes are the ones a live benchmark needs: a
\emph{corporate-action and event axis} (earnings, dividends, splits, M\&A,
defaults), whose events arrive daily and drive freshness, and an \emph{entity
layer} (issuers, intermediaries, funds, infrastructure, regulators, key
persons) that lists the kinds of entity a question can be keyed on. The
taxonomy fixes the benchmark's scope. Our questions are drawn from the domains
that the data source serves well: listed equities, funds, indices, macro
series, market microstructure, and corporate events. Derivatives, FX,
commodities, insurance, and structured credit lie outside the current scope.

\subsection{Stage 1: Answer-First Construction}
\label{sec:answerfirst}

The pipeline calls tools first, observes real responses, and only then derives
questions whose answers are executed values, so the model never writes a gold
number.

\paragraph{Probe and coverage.}
For each asset class (A-share, Hong Kong and US equities, indices,
exchange-traded and off-exchange funds), a probe samples entities and marks a
tool as usable for the class when a strict majority of them return data;
entity-free tools such as macro series are probed once. Coverage then draws,
for every usable tool, an entity at each of eleven web-visibility levels,
moving to another entity at the same level whenever a call returns no data, so
every kept call is backed by a tool that actually serves its entity. An
entity's web-visibility index (0--10) counts the web search results for its
name that mention it, which spreads the pool from obscure to well-known
instruments instead of concentrating it on popular ones.

\paragraph{Program enumeration and filtering.}
For each observed response, the pipeline enumerates every extractor that the
extractor language can execute to a non-null value: \texttt{latest} over each
rolling field of a dated series, \texttt{match} on an identity key for
cross-sectional tables, and \texttt{snapshot} for single-record responses.
Mechanical filters reject metadata, pagination, and display fields, result-set
counts, fields of tools whose data are static (a value that never changes
cannot drift), and \texttt{latest} selectors whose newest period holds
conflicting values. A language model then reviews each remaining target for
whether it is worth asking. The verdict is keyed by a \emph{semantic
signature}, namely the tool, its non-entity arguments, the extractor structure,
and the refresh cadence, and is shared by every entity with that signature.
Question diversity is thus a property of the enumeration, not of model
behaviour.

\paragraph{Verbalization.}
For each kept signature, one language-model call sees an observed response
together with the fixed extractor and writes the question text, a one-sentence
conclusion, the time anchor, the rendering format, and a proposed matching
rule; it cannot change which row or field is selected. The gold answer is
computed by executing the extractor. A candidate is kept only if (i)~the
executed value agrees with the value the model states, within $1\%$ for numeric
answers; (ii)~the rendered unit agrees with the tool's own display string for
that field; and (iii)~the text contains no absolute date, no reference to
hidden tool context, and no tool or field name. Rejected candidates are retried
with the rejection reason as feedback. The accepted template is then bound to
every other entity with the same signature by substituting names and codes, and
the checks run again on each binding. A final pass rewrites ambiguous phrasings
and keeps the original text for audit, and a validator checks that the
question text, the call arguments, and the extractor's row filter all name the
same entity; refresh and evaluation repeat this check. The resulting inventory
holds 3{,}320 questions over 34 tools. An optional tool-necessity gate, in
which a tool-using agent must succeed over $k{=}3$ runs while a model without
tools must fail, can further tighten the inventory when a stricter pool is
desired.

\subsection{Stage 2: Pool Selection and Daily Refresh}
\label{sec:daily}

\paragraph{Pool selection.}
The evaluation pool is drawn from the daily-cadence part of the inventory
without any language-model call. Signatures that every model answered correctly
in a preliminary five-model run are dropped. Each remaining signature is
re-instantiated on fresh entities spread over the eleven web-visibility levels:
each new entity is probed once (one call serves every signature that shares
it), the signature's extractor is reused unchanged, and the question text is
the signature's template with the entity substituted. Entities are assigned so
that a signature appears in both a test and a training split while no entity
appears in both. Entity-free macro questions are reused as they are.

\paragraph{Daily refresh.}
The daily job re-executes every program in the pool. It rebuilds each call's
time window from the benchmark date rather than the machine clock, changing
only date arguments, and anchors US equities to the last settled session
rather than an unsettled current-session bar. Scoring a past run always uses
the frozen snapshot of its date rather than re-querying historical data.
It then runs the program through the shared execution layer, applies the
extractor, and writes the dated snapshot. A question whose entity has no data
that day (for instance, no block trades in the window) is left out of that
day's served set but stays in the frozen pool, because event-driven data can
return later. How often a question's answer flips and how often it fails to
resolve are reportable properties of the benchmark; questions that fail
repeatedly are candidates for retirement. Section~\ref{sec:drift} measures the
flip rate.

\section{Experiments}
\label{sec:results}

\subsection{Experimental Setup}
\label{sec:setup}

\paragraph{Temporal-analysis cohort.}
\begin{sloppypar}
The frozen pool used for the temporal analyses
(Sections~\ref{sec:drift} and~\ref{sec:temporal}) has $153$ questions,
minted from the inventory on 2026-08-05: $13$ at each of the eleven
web-visibility levels plus $10$
entity-free macro questions, over $12$ tools. Every question is single-hop (one
tool call plus an extractor) and in Chinese, about A-share, Hong Kong, and US
instruments, funds, indices, and macro and rate series. Each day the served set
is the questions that resolve: about $110$ per day, and $109$ on 2026-09-16, the
snapshot used for the temporal-drift analysis. Of these, $106$ ($97\%$) have numeric answers
and $3$ have string answers; all use the \texttt{approx} rule.
\end{sloppypar}

\paragraph{Agent setup.}
For the temporal and failure analyses, the agent runs in a ReAct-style
tool-calling loop~\cite{react} with two generic web tools, web search and page
fetch, and at most $15$ turns. Its system prompt gives the benchmark date,
which the relative time anchors require,
and instructs the agent to use tool-returned evidence rather than parametric
memory. The episode ends
when the agent replies without a tool call; at the turn limit the tools are
withdrawn and the agent must give its best answer. Scoring reads this final
reply, and an empty reply scores $0$. The professional data source is not
exposed to the agent: it only executes the gold programs, so the evaluation
measures whether an agent with ordinary web search can recover a value
established from a professional source. Because the question text is
tool-agnostic, giving the agent the professional tools on the same questions
would measure how far web search substitutes for professional data; we leave
that condition to future work.

\paragraph{Models and metric.}
We evaluate six contemporary tool-calling models (Table~\ref{tab:singlehop});
parenthesized labels denote each model's reasoning configuration.
Our evaluation harness scores \textbf{answer accuracy}: a binary ($0/1$)
judgment per question against the single gold value under the question's
matching rule~$\tau$ (Section~\ref{sec:design}), averaged over questions.
The judge is \texttt{gemini-3.5-flash} at temperature $0$, fixed across agents
within a run and drawn from a model family absent from the leaderboard. It
receives $\tau$ by name, never as a number, and judges only
the final committed value, not values mentioned along the way.

\subsection{Main Results}
\label{sec:main}

Table~\ref{tab:singlehop} reports six models under the \emph{search-only}
condition, where DeepSeek V4.1 Flash achieves the highest score ($76.7$), followed
by Claude Opus 5 (Max) at $76.2$. Step 5 Preview (High) and GPT-6 Astra (Max)
both score $74.5$; GLM-5.3 (Max) and Kimi K3 (Max) score $73.3$ and $70.9$,
respectively. Even the strongest model misses nearly a quarter of the
questions, leaving clear headroom for search agents that must recover
professionally defined values from the open web. The following sections
examine how gold answers move over time, how per-question correctness turns
over, which questions discriminate, and where agents fail, each on the
snapshots specified in its setup.

\begin{table}[t]
\caption{\textbf{Leaderboard.} Answer accuracy (\%) of six tool-calling models
under the \emph{search-only} condition (no professional financial tools).
Parenthesized labels denote reasoning configuration.}
\label{tab:singlehop}
\centering
\begin{tabular}{@{}lr@{}}
\toprule
Model & Accuracy (\%) \\
\midrule
DeepSeek V4.1 Flash       & \textbf{76.7} \\
Claude Opus 5 (Max)       & 76.2 \\
Step 5 Preview (High)     & 74.5 \\
GPT-6 Astra (Max)         & 74.5 \\
GLM-5.3 (Max)             & 73.3 \\
Kimi K3 (Max)             & 70.9 \\
\bottomrule
\end{tabular}
\end{table}

\subsection{Gold Answers Drift Day over Day}
\label{sec:drift}

\begin{figure}[t]
\centering
\includegraphics[width=\columnwidth]{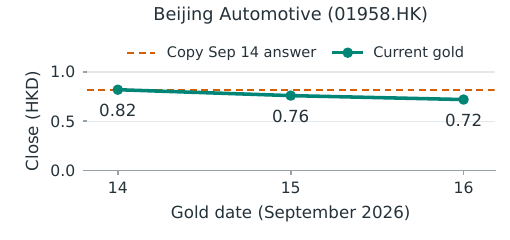}
\caption{\textbf{One question, three gold dates.} The question asks for the
latest-trading-day close (Chinese; glossed) and is byte-identical across dates.
Only the program's date window changes. The dashed line copies September 14's
gold: it matches on that date but not on the next two.}
\Description{Line plot of Beijing Automotive's latest-trading-day close on
September 14, 15, and 16, 2026: 0.82, 0.76, and 0.72 HKD. A dashed line holds
the September 14 value at 0.82. The vertical axis starts at zero.}
\label{fig:drift-example}
\end{figure}

\paragraph{Fixed questions, moving values.}
We audit the dated snapshots from 2026-09-14 to 2026-09-16; one example is shown
in Figure~\ref{fig:drift-example}. They serve $110$, $110$, and $109$ questions,
respectively; all gold comparisons below use the
same $109$ qids that resolve on every date. For each qid, we verify that the
question text agrees with the frozen record, the extractor is fixed, and the
executed non-date arguments remain unchanged. Date arguments, including the
last-settled-session window for US equities, are the only varying inputs.

\begin{figure*}[t]
\centering
\includegraphics[width=\textwidth]{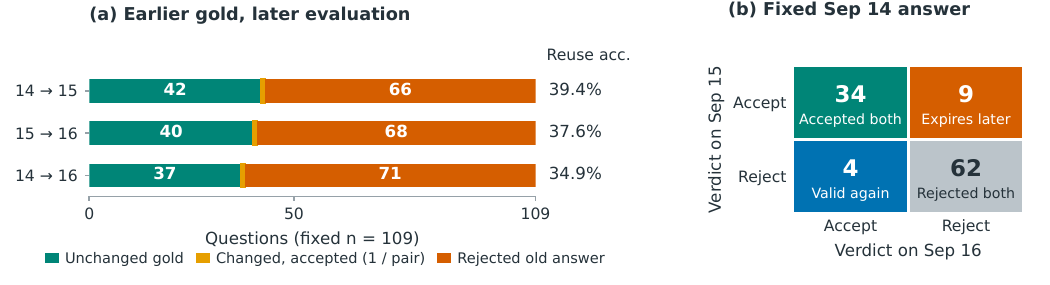}
\caption{\textbf{Literal answer reuse: failure and return to validity.}
Gold dates are in September 2026; both panels use the same $109$ questions.
(a) Earlier gold is copied into each later evaluation: unchanged values pass,
one changed value per pair still passes, and the rest are rejected under the
fixed rule. (b) The September 14 answer is held fixed at both later dates;
four answers fail on September 15 and pass again on September 16.}
\Description{Left: stacked bars for September 14 to 15, 15 to 16, and 14 to 16.
Unchanged values number 42, 40, and 37; each comparison accepts one changed
value and rejects 66, 68, and 71 old answers. Reuse accuracies are 39.4, 37.6,
and 34.9 percent. Right: a two-by-two matrix of the fixed September 14 answer's
validity on September 15 and 16: 34 accepted both, 9 accepted then rejected,
4 rejected then accepted, and 62 rejected both.}
\label{fig:snapshot-reuse}
\end{figure*}

Of these $109$ questions, $76$ ($70\%$) change their formatted gold value at
least once: $67$ and $69$ between consecutive days, and $72$ across the span
(Figure~\ref{fig:snapshot-reuse}). The paths are not uniformly monotone: $56$
questions take three distinct values, while $4$ return to their first-day value
after changing on the middle date. The remaining $33$ retain the same formatted
answer. Refreshing a program therefore does not require its output to change;
a slow-moving rate or a repeated market value can legitimately persist.
Drift is concentrated in market-linked fields: \texttt{get\_hk\_stock\_kline}
$45/54$, \texttt{get\_us\_stock\_kline} $18/18$, \texttt{get\_kline} $4/4$,
and \texttt{get\_valuation\_metrics\_daily} $4/5$ change at least once,
whereas macro-rate answers barely move. Nonzero signs reverse on $5$ and $8$
questions at the one-day transitions ($9$ across the span), a change that no
numeric precision allowance can turn into an equivalent direction.

\begin{figure}[t]
\centering
\includegraphics[width=\columnwidth]{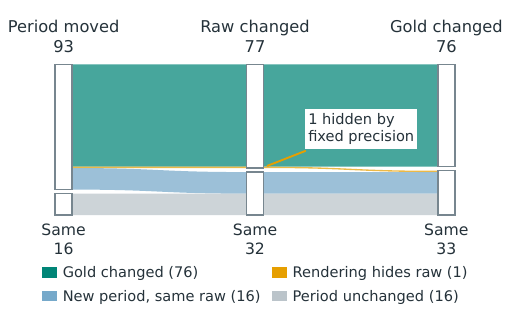}
\caption{\textbf{From new periods to new answers.} All $109$ questions over
September 14--16, 2026. Band widths count questions and trace the joint paths
from reported observation periods to raw values and fixed rendered gold.
Sixteen questions acquire a new period without a new raw value; one raw change
is hidden by rendering.}
\Description{Three-stage flow diagram. Reported periods move for 93 questions
and remain the same for 16. Raw values change for 77 and stay the same for 32.
Rendered gold changes for 76 and stays the same for 33. Four paths have counts
76, 1, 16, and 16; the one-question rendering loss is called out explicitly.}
\label{fig:temporal-layers}
\end{figure}

\paragraph{Three clocks, not one.}
The benchmark date, the request's cutoff, and the provider-reported observation
period need not coincide. Across the common panel, observed periods advance
at least once for $93$ questions and stay fixed for $16$
(Figure~\ref{fig:temporal-layers}). All $76$ formatted-gold changes occur in the
first group, but $17$ questions in that group retain their formatted answer
even as their reported period advances. Thus the $33$ unchanged answers
combine repeated values at new periods with values from unchanged periods;
an unchanged answer alone is not evidence that the refresh failed.
Because each snapshot retains both the executed arguments and the observed
period, every such case remains inspectable.

The distinction also explains what ``latest'' means operationally. For all
$18$ US-equity programs, the request's \texttt{end\_date} is September 11, 14,
and 15 for benchmark dates September 14, 15, and 16, respectively, reflecting
the last settled session. For Hong Kong stock 08033.HK, the volume changes from
$372{,}000$ to $50{,}000$ shares and then repeats; the last two snapshots both
report September 15 as the observed period even though the request window
advances to September 16. In contrast, an RMB deposit-rate
program retains $1.30\%$ and a January 15 observation throughout, while its
\texttt{query\_date} advances daily. These are different time anchors, not a
requirement to return a number bearing the benchmark date. An agent that
substitutes an unsettled quote or a zero-volume calendar-day row can therefore
answer a different question from the one the program defines.

\paragraph{Rendering is part of the measurement.}
The raw extracted value changes at least once on $77$ questions, compared with
$76$ after formatting. The extra case is stock 00361.HK: its raw opening price
moves from $0.039$ to $0.038$ HKD, but all three snapshots record $0.04$ HKD.
At the first one-day transition, $69$ raw values change but only $67$
formatted values do; at the endpoints the corresponding counts are $74$ and
$72$. The drift statistic therefore depends on the published precision, not
just the underlying series. We keep that precision fixed rather than changing
it after seeing the data, and distinguish a raw-value movement, a changed
rendered gold, and a rejected old answer under $\tau$. The last of these is
the relevant test of whether a copied benchmark answer still receives credit.

\paragraph{Does a once-correct answer still pass?}
A change in the formatted gold is not automatically a failed answer under
$\tau$. We therefore evaluate a \emph{literal snapshot-reuse diagnostic}:
for each fixed qid, submit its earlier formatted gold as the prediction against
the later snapshot. Identical values pass by the matching rule; all changed
values are judged with the canonical prompt, \texttt{gemini-3.5-flash} at
temperature $0$, and the question's fixed rule. This requires no search,
agent execution, or gold refresh. Reusing the preceding snapshot fails on
$66/109$ and $68/109$ questions at the two one-day transitions; reusing the
first snapshot on the third day fails on $71/109$. One changed value per
comparison is still accepted, illustrating why exact gold-string changes
must be distinguished from tolerance-aware failure. Thus, a literal leaked
snapshot loses validity on a majority of served questions within a day. The
diagnostic measures the shelf life of a copied answer directly, independent of
any particular model.

\paragraph{Validity can return, not just decay.}
Holding the September 14 answer fixed and checking it against both later
snapshots produces four paths: $34$ answers pass on both later dates, $62$
fail on both, $9$ pass on September 15 but fail on September 16, and $4$ fail
on September 15 but pass again on September 16. Of these four returns to
validity, three also return to the initial raw value; one returns only at the
published precision. For stock 01022.HK, the raw back-adjusted price change
is $-0.01$, $0.005$, and $-0.005$ HKD, rendered as $-0.01$, $0.01$, and
$-0.01$ HKD. The old negative answer contradicts the middle day's direction
but matches the third day's rendered gold.
A per-qid trajectory captures temporary invalidity and later equivalence,
which a single aggregate ``freshness'' number cannot distinguish.

\subsection{Per-Question Correctness Turnover}
\label{sec:temporal}
We further track archived single-run trajectories of three models,
DeepSeek V4 Pro (0813), Kimi K3, and GLM-5.3 (agent configurations separate
from Table~\ref{tab:singlehop}), each scored against its own date's gold
snapshot. We compare the September 14 and 16 endpoints; September 15 is
omitted because the page-fetch tool was unavailable during that day's runs.
The three models share $85$ qids with
identical question text at both endpoints. We recompute their scores on this
fixed panel (Table~\ref{tab:temporal-agents}); $56$ of its gold values differ
between the endpoints. DeepSeek's accuracy is unchanged at $57/85$, yet $11$
previously correct questions become wrong and $11$ previously wrong questions
become correct. Kimi and GLM also exchange correct and incorrect items.
Each model is judged correct at both endpoints on $31$ of the $56$
gold-changing questions, while some verdicts also change on questions with
identical endpoint gold. These are distinct signals: movement of the reference
value and variation in recovering it. Preserving qids makes both visible,
whereas a single aggregate score conceals the turnover.

\begin{table}[t]
\caption{\textbf{Paired-agent panel.} The same $85$ questions
against gold snapshots dated September 14 and 16, 2026. $1\to0$ and $0\to1$
count changes in per-question
correctness against each date's own gold.}
\label{tab:temporal-agents}
\centering\small
\begin{tabular}{@{}lrrrr@{}}
\toprule
Model & 14 (\%) & 16 (\%) & $1\to0$ & $0\to1$ \\
\midrule
DeepSeek V4 Pro (0813) & 67.1 & 67.1 & 11 & 11 \\
Kimi K3               & 70.6 & 67.1 & 11 & 8 \\
GLM-5.3               & 69.4 & 67.1 & 12 & 10 \\
\bottomrule
\end{tabular}
\end{table}

\begin{figure}[t]
\centering
\includegraphics[width=\columnwidth]{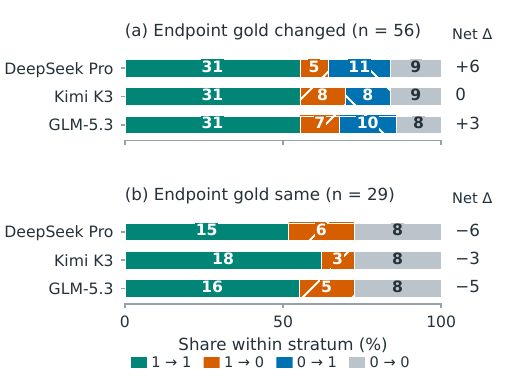}
\caption{\textbf{Paired correctness by endpoint gold movement.} The same
$85$ questions as Table~\ref{tab:temporal-agents}, split by whether September
14 and 16 gold differs. Bar lengths are normalized within each stratum;
segment labels are question counts, and ``Net $\Delta$'' is the change in the
number correct. Unlike net accuracy, the four transitions retain both gains
and losses.}
\Description{Two groups of stacked bars for DeepSeek Pro, Kimi K3, and
GLM-5.3. In the 56 changed-gold questions, both-correct, correct-to-wrong,
wrong-to-correct, and both-wrong counts are 31/5/11/9, 31/8/8/9, and 31/7/10/8.
In the 29 unchanged-endpoint-gold questions, the counts are 15/6/0/8,
18/3/0/8, and 16/5/0/8. Net correct-count changes are +6/0/+3 and -6/-3/-5,
respectively.}
\label{fig:agent-turnover}
\end{figure}

\paragraph{Condition on whether the answer changed.}
Figure~\ref{fig:agent-turnover} partitions the same panel into $56$ questions
whose formatted gold differs between endpoints and $29$ whose endpoint gold
agrees. Within the changed-gold group, DeepSeek gains $11$ correct items and
loses $5$, increasing from $36/56$ to $42/56$; Kimi gains and loses $8$,
remaining at $39/56$; GLM gains $10$ and loses $7$, increasing from $38/56$ to
$41/56$. On the unchanged-endpoint group, however, the models lose $6$, $3$,
and $5$ correct items, respectively, with no wrong-to-correct transitions.
DeepSeek's six-item gain on changing gold is exactly offset by a six-item loss
on unchanged gold. Kimi's three-item overall decline occurs entirely in the
unchanged group, and GLM's three-item gain in the changed group is outweighed
by a five-item loss in the unchanged group. Aggregate differences cannot
therefore be read as counts of failures caused by a moving gold answer.
``Unchanged'' here means identical endpoint gold, not an unchanged raw series
or an answer that stayed constant on the intervening date.

For binary scores, let $L$ count correct-to-wrong transitions and $G$ count
wrong-to-correct transitions. The change in the number correct is $G-L$,
whereas the number of changed verdicts is $G+L$. These logs have $22$, $19$,
and $22$ changed verdicts for DeepSeek, Kimi, and GLM, despite net changes of
$0$, $-3$, and $-2$. Two-date correctness is a separate criterion: only $46$,
$49$, and $47$ of the $85$ questions are correct at both endpoints,
respectively, and each model has $31$ such cases in the changed-gold group.
Reporting the paired transitions therefore adds information beyond either day's
accuracy. We use this panel descriptively, to show what per-question tracking
reveals; the same protocol directly supports controlled longitudinal
comparisons that hold the agent configuration fixed across dates.

\subsection{Which Questions Discriminate?}
\label{sec:discrim}

Not every question separates models. In an $86$-question pilot,
answered by $11$ models with web search and page fetch
under a fixed judge, $41$ questions were solved by every model and $10$ by none,
leaving $35$ that split the field. The discriminating questions concentrate on
\emph{confusable financial calibers} and \emph{edge-market microstructure}
(e.g.\ return on equity via \texttt{get\_financial\_index}, block-trade detail,
Hong Kong and US K-lines), whereas plain A-share balance-sheet and K-line
questions are solved by nearly every model. Difficulty thus sits where the
design places it, in resolving the right field and caliber among near-synonyms.
Pool selection (Section~\ref{sec:daily}) uses the same signal by dropping
signatures that every model solves.

\subsection{Failure Analysis}
\label{sec:failure}

Evaluation trajectories show that errors rarely come from a lack of
searching: agents typically issue several searches and cross-check two or
three pages before answering. The errors come instead from mismatches between
what public pages present and what the question specifies. \emph{(i) Precisely
defined calibers.} Back-adjusted prices of Hong Kong stocks depend on the
adjustment method, so a value read from a portal that adjusts differently, or
recomputed from a split history, misses the gold; for counts of
moving-average crossings, agents compute their own count under their own
definition; and for the daily trading value of US stocks, agents approximate
it as volume times closing price instead of retrieving the reported value.
\emph{(ii) A different ``latest'' session.} For an illiquid Hong Kong stock,
the latest trading session is the most recent one with trades, but agents
report the zero-volume calendar day shown on public pages. \emph{(iii)
Unverified secondary values.} Agents sometimes report a third-party quote
without checking the issuing source; for the two-year Chinese government bond
yield, agents that used an aggregator's figure ($1.247\%$) missed the official
curve value ($1.2439\%$) that other agents found. Under web search, then, the
hard part is matching the exact caliber and period that the question defines.

\section{Conclusion}
\label{sec:conclusion}

LiveSearch turns a financial search benchmark from a fixed test set into a
live measurement instrument by storing each question as executable verifier
code over professional financial data tools. The question text is frozen and
the gold answer is the program's live output, so automated curation and a
daily refresh without any language model follow from one representation, and
the model verbalizes but never decides the answer. Under general web search,
six contemporary models reach $70.9\%$--$76.7\%$ answer accuracy. Over three
consecutive days, $70\%$ of served questions change their gold value, and
literal snapshot reuse fails on a majority of questions at both one-day
transitions even under the fixed matching rules. Per-question tracking further
shows that aggregate accuracy can stay flat while individual verdicts turn
over, a signal that only stable question identities and dated gold snapshots
expose. The approach is not specific to finance: any domain with a stable
programmatic data interface can host a live, auto-refreshing benchmark built
the same way.

\section*{Limitations}

The data source serves equities, funds, indices, macro, microstructure, and
corporate events well, and derivatives, FX, commodities, insurance, and
structured credit weakly or not at all, so our
claims about professional-source grounding are limited to the covered domains.
The data source is also a proprietary service that external researchers cannot
call freely. We mitigate this with documented tool input/output schemas, frozen
timestamped gold snapshots (the shareable (question, timestamp, value) tuples),
and gated evaluation access; live re-execution requires the service. Gold
values also inherit the accuracy of the upstream professional source.

Scoring also depends on an LLM judge. Fixing its configuration within a run
and using a model family absent from the leaderboard is intended to limit
bias, but does not eliminate judge error or the position and verbosity effects
known in LLM-based evaluation~\cite{llmjudge}. 

\bibliographystyle{ACM-Reference-Format}
\bibliography{references}

@misc{fingaia,
  title = {{FinGAIA}: A Chinese Benchmark for {AI} Agents in Real-World Financial Domain},
  author = {Lingfeng Zeng and Fangqi Lou and Zixuan Wang and Jiajie Xu and Jinyi Niu and Mengping Li and Yifan Dong and Qi Qi and Wei Zhang and Ziwei Yang and Jun Han and Ruilun Feng and Ruiqi Hu and Lejie Zhang and Zhengbo Feng and Yicheng Ren and Xin Guo and Zhaowei Liu and Dongpo Cheng and Weige Cai and Liwen Zhang},
  year = {2025},
  note = {arXiv:2507.17186},
  url = {https://arxiv.org/abs/2507.17186}
}

@misc{finsearchcomp,
  title = {{FinSearchComp}: Towards a Realistic, Expert-Level Evaluation of Financial Search and Reasoning},
  author = {Liang Hu and Jianpeng Jiao and Jiashuo Liu and Yanle Ren and Zhoufutu Wen and Kaiyuan Zhang and Xuanliang Zhang and Xiang Gao and Tianci He and Fei Hu and Yali Liao and Zaiyuan Wang and Chenghao Yang and Qianyu Yang and Mingren Yin and Zhiyuan Zeng and Ge Zhang and Xinyi Zhang and Xiying Zhao and Zhenwei Zhu and Hongseok Namkoong and Wenhao Huang and Yuwen Tang},
  year = {2025},
  note = {arXiv:2509.13160},
  url = {https://arxiv.org/abs/2509.13160}
}

@misc{mcpatlas,
  title = {{MCP-Atlas}: A Large-Scale Benchmark for Tool-Use Competency with Real {MCP} Servers},
  author = {Chaithanya Bandi and Razvan-Gabriel Dumitru and Ben Hertzberg and Divyansh Agarwal and Geobio Boo and Tejas Polakam and Sami Hassaan and Jeff Da and HiJae Kim and Vipul Gupta and Manasi Sharma and Andrew Park and Martin Dimakis and Ernesto Gabriel Hernandez Montoya and Dan Rambado and Ivan Salazar and Rafael Cruz and MohammadHossein Rezaei and Chetan Rane and Ben Levin and Daniel Yue Zhang and Brad Kenstler and Bing Liu},
  year = {2026},
  note = {arXiv:2602.00933},
  url = {https://arxiv.org/abs/2602.00933}
}

@misc{livemcp,
  title = {{LiveMCPBench}: Can Agents Navigate an Ocean of {MCP} Tools?},
  author = {Guozhao Mo and Wenliang Zhong and Jiawei Chen and Qianhao Yuan and Xuanang Chen and Yaojie Lu and Hongyu Lin and Ben He and Xianpei Han and Le Sun},
  year = {2026},
  note = {arXiv:2508.01780},
  url = {https://arxiv.org/abs/2508.01780}
}

@misc{financeagentbench,
  title = {Finance Agent Benchmark: Benchmarking {LLM}s on Real-World Financial Research Tasks},
  author = {Antoine Bigeard and Langston Nashold and Rayan Krishnan and Shirley Wu},
  year = {2025},
  note = {arXiv:2508.00828},
  url = {https://arxiv.org/abs/2508.00828}
}

@misc{sun2025finresearchbenchlogictreebased,
  title = {{FinResearchBench}: A Logic Tree Based Agent-as-a-Judge Evaluation Framework for Financial Research Agents},
  author = {Rui Sun and Zuo Bai and Wentao Zhang and Yuxiang Zhang and Li Zhao and Shan Sun and Zhengwen Qiu},
  year = {2025},
  note = {arXiv:2507.16248},
  url = {https://arxiv.org/abs/2507.16248}
}

@misc{luan2026finresearchbenchiideepresearch,
  title = {{FinResearchBench II}: A Deep Research Benchmark with Consensus-Derived Gold Rubrics for Distinguishing Financial Report Quality},
  author = {Beidi Luan and Rui Sun and Sinuo Wang and Yan Gu and Chao Li and Zhenliang Xiong and Jing Li and Zuo Bai},
  year = {2026},
  note = {arXiv:2607.12252},
  url = {https://arxiv.org/abs/2607.12252}
}

@misc{webdancer,
  title = {{WebDancer}: Towards Autonomous Information Seeking Agency},
  author = {Jialong Wu and Baixuan Li and Runnan Fang and Wenbiao Yin and Liwen Zhang and Zhengwei Tao and Dingchu Zhang and Zekun Xi and Gang Fu and Yong Jiang and Pengjun Xie and Fei Huang and Jingren Zhou},
  year = {2025},
  note = {arXiv:2505.22648},
  url = {https://arxiv.org/abs/2505.22648}
}

@misc{websailor,
  title = {{WebSailor}: Navigating Super-Human Reasoning for Web Agents},
  author = {Kuan Li and Zhongwang Zhang and Huifeng Yin and Liwen Zhang and Litu Ou and Jialong Wu and Wenbiao Yin and Baixuan Li and Zhengwei Tao and Xinyu Wang and Weizhou Shen and Junkai Zhang and Dingchu Zhang and Xixi Wu and Yong Jiang and Ming Yan and Pengjun Xie and Fei Huang and Jingren Zhou},
  year = {2025},
  note = {arXiv:2507.02592},
  url = {https://arxiv.org/abs/2507.02592}
}

@misc{webwalker,
  title = {{WebWalker}: Benchmarking {LLM}s in Web Traversal},
  author = {Jialong Wu and Wenbiao Yin and Yong Jiang and Zhenglin Wang and Zekun Xi and Runnan Fang and Linhai Zhang and Yulan He and Deyu Zhou and Pengjun Xie and Fei Huang},
  year = {2025},
  note = {arXiv:2501.07572},
  url = {https://arxiv.org/abs/2501.07572}
}

@inproceedings{react,
  title = {{ReAct}: Synergizing Reasoning and Acting in Language Models},
  author = {Shunyu Yao and Jeffrey Zhao and Dian Yu and Nan Du and Izhak Shafran and Karthik Narasimhan and Yuan Cao},
  booktitle = {International Conference on Learning Representations},
  year = {2023},
  url = {https://arxiv.org/abs/2210.03629}
}

@inproceedings{toolllm,
  title = {{ToolLLM}: Facilitating Large Language Models to Master 16,000+ Real-World {API}s},
  author = {Yujia Qin and Shihao Liang and Yining Ye and Kunlun Zhu and Lan Yan and Yaxi Lu and Yankai Lin and Xin Cong and Xiangru Tang and Bill Qian and Sihan Zhao and Lauren Hong and Runchu Tian and Ruobing Xie and Jie Zhou and Mark Gerstein and Dahai Li and Zhiyuan Liu and Maosong Sun},
  booktitle = {International Conference on Learning Representations},
  year = {2024},
  url = {https://arxiv.org/abs/2307.16789}
}

@inproceedings{webarena,
  title = {{WebArena}: A Realistic Web Environment for Building Autonomous Agents},
  author = {Shuyan Zhou and Frank F. Xu and Hao Zhu and Xuhui Zhou and Robert Lo and Abishek Sridhar and Xianyi Cheng and Tianyue Ou and Yonatan Bisk and Daniel Fried and Uri Alon and Graham Neubig},
  booktitle = {International Conference on Learning Representations},
  year = {2024},
  url = {https://arxiv.org/abs/2307.13854}
}

@inproceedings{finqa,
  title = {{FinQA}: A Dataset of Numerical Reasoning over Financial Data},
  author = {Zhiyu Chen and Wenhu Chen and Charese Smiley and Sameena Shah and Iana Borova and Dylan Langdon and Reema Moussa and Matt Beane and Ting-Hao Huang and Bryan Routledge and William Yang Wang},
  booktitle = {Proceedings of the 2021 Conference on Empirical Methods in Natural Language Processing},
  pages = {3697--3711},
  year = {2021},
  url = {https://aclanthology.org/2021.emnlp-main.300}
}

@inproceedings{convfinqa,
  title = {{ConvFinQA}: Exploring the Chain of Numerical Reasoning in Conversational Finance Question Answering},
  author = {Zhiyu Chen and Shiyang Li and Charese Smiley and Zhiqiang Ma and Sameena Shah and William Yang Wang},
  booktitle = {Proceedings of the 2022 Conference on Empirical Methods in Natural Language Processing},
  pages = {6279--6292},
  year = {2022},
  url = {https://aclanthology.org/2022.emnlp-main.421}
}

@misc{financebench,
  title = {{FinanceBench}: A New Benchmark for Financial Question Answering},
  author = {Pranab Islam and Anand Kannappan and Douwe Kiela and Rebecca Qian and Nino Scherrer and Bertie Vidgen},
  year = {2023},
  note = {arXiv:2311.11944},
  url = {https://arxiv.org/abs/2311.11944}
}

@misc{bloomberggpt,
  title = {{BloombergGPT}: A Large Language Model for Finance},
  author = {Shijie Wu and Ozan Irsoy and Steven Lu and Vadim Dabravolski and Mark Dredze and Sebastian Gehrmann and Prabhanjan Kambadur and David Rosenberg and Gideon Mann},
  year = {2023},
  note = {arXiv:2303.17564},
  url = {https://arxiv.org/abs/2303.17564}
}

@misc{pixiu,
  title = {{PIXIU}: A Large Language Model, Instruction Data and Evaluation Benchmark for Finance},
  author = {Qianqian Xie and Weiguang Han and Xiao Zhang and Yanzhao Lai and Min Peng and Alejandro Lopez-Lira and Jimin Huang},
  year = {2023},
  note = {arXiv:2306.05443},
  url = {https://arxiv.org/abs/2306.05443}
}

@inproceedings{finben,
  title = {{FinBen}: A Holistic Financial Benchmark for Large Language Models},
  author = {Qianqian Xie and Weiguang Han and Zhengyu Chen and Ruoyu Xiang and Xiao Zhang and Yueru He and Mengxi Xiao and Dong Li and Yongfu Dai and Duanyu Feng and Yijing Xu and Haoqiang Kang and Ziyan Kuang and Chenhan Yuan and Kailai Yang and Zheheng Luo and Tianlin Zhang and Zhiwei Liu and Guojun Xiong and Zhiyang Deng and Yuechen Jiang and Zhiyuan Yao and Haohang Li and Yangyang Yu and Gang Hu and Jiajia Huang and Xiao-Yang Liu and Alejandro Lopez-Lira and Benyou Wang and Yanzhao Lai and Hao Wang and Min Peng and Sophia Ananiadou and Jimin Huang},
  booktitle = {Advances in Neural Information Processing Systems},
  volume = {37},
  year = {2024},
  url = {https://arxiv.org/abs/2402.12659}
}

@misc{fintoolbench,
  title = {{FinToolBench}: Evaluating {LLM} Agents for Real-World Financial Tool Use},
  author = {Jiaxuan Lu and Kong Wang and Yemin Wang and Qingmei Tang and Hongwei Zeng and Xiang Chen and Jiahao Pi and Shujian Deng and Lingzhi Chen and Yi Fu and Kehua Yang and Xiao Sun},
  year = {2026},
  note = {arXiv:2603.08262},
  url = {https://arxiv.org/abs/2603.08262}
}

@misc{frontierfinance,
  title = {{FrontierFinance}: A Challenging Benchmark for Measuring Frontier Intelligence of Finance Agents},
  author = {Yuhao Zhang and O. Ozan Koyluoglu and Thejas Venkatesh and Richard Diehl Martinez and Vishank Bhatia and Arash Alidoust and Ashwin Paranjape},
  year = {2026},
  note = {arXiv:2608.11683},
  url = {https://arxiv.org/abs/2608.11683}
}

@misc{fintrace,
  title = {{FinTrace}: Holistic Trajectory-Level Evaluation of {LLM} Tool Calling for Long-Horizon Financial Tasks},
  author = {Yupeng Cao and Haohang Li and Weijin Liu and Wenbo Cao and Anke Xu and Lingfei Qian and Xueqing Peng and Minxue Tang and Zhiyuan Yao and Jimin Huang and K. P. Subbalakshmi and Zining Zhu and Jordan W. Suchow and Yangyang Yu},
  year = {2026},
  note = {arXiv:2604.10015},
  url = {https://arxiv.org/abs/2604.10015}
}

@inproceedings{livebench,
  title = {{LiveBench}: A Challenging, Contamination-Limited {LLM} Benchmark},
  author = {Colin White and Samuel Dooley and Manley Roberts and Arka Pal and Ben Feuer and Siddhartha Jain and Ravid Shwartz-Ziv and Neel Jain and Khalid Saifullah and Sreemanti Dey and Shubh Agrawal and Sandeep Singh Sandha and Siddartha Naidu and Chinmay Hegde and Yann LeCun and Tom Goldstein and Willie Neiswanger and Micah Goldblum},
  booktitle = {International Conference on Learning Representations},
  year = {2025},
  url = {https://arxiv.org/abs/2406.19314}
}

@misc{evobrowsecomp,
  title = {{EvoBrowseComp}: Benchmarking Search Agents on Evolving Knowledge},
  author = {Yunhan Wang and Jiaan Wang and Lianzhe Huang and Xianfeng Zeng and Fandong Meng},
  year = {2026},
  note = {arXiv:2606.13120},
  url = {https://arxiv.org/abs/2606.13120}
}

@inproceedings{qgsql,
  title = {Question Generation from {SQL} Queries Improves Neural Semantic Parsing},
  author = {Daya Guo and Yibo Sun and Duyu Tang and Nan Duan and Jian Yin and Hong Chi and James Cao and Peng Chen and Ming Zhou},
  booktitle = {Proceedings of the 2018 Conference on Empirical Methods in Natural Language Processing},
  pages = {1597--1607},
  year = {2018},
  url = {https://aclanthology.org/D18-1180}
}

@inproceedings{autoqa,
  title = {{AutoQA}: From Databases to {QA} Semantic Parsers with Only Synthetic Training Data},
  author = {Silei Xu and Sina J. Semnani and Giovanni Campagna and Monica S. Lam},
  booktitle = {Proceedings of the 2020 Conference on Empirical Methods in Natural Language Processing},
  pages = {422--434},
  year = {2020},
  url = {https://arxiv.org/abs/2010.04806}
}

@inproceedings{llmjudge,
  title = {Judging {LLM}-as-a-Judge with {MT-Bench} and {Chatbot Arena}},
  author = {Lianmin Zheng and Wei-Lin Chiang and Ying Sheng and Siyuan Zhuang and Zhanghao Wu and Yonghao Zhuang and Zi Lin and Zhuohan Li and Dacheng Li and Eric P. Xing and Hao Zhang and Joseph E. Gonzalez and Ion Stoica},
  booktitle = {Advances in Neural Information Processing Systems},
  volume = {36},
  year = {2023},
  url = {https://arxiv.org/abs/2306.05685}
}

@misc{mcp,
  title = {Introducing the Model Context Protocol},
  author = {{Anthropic}},
  year = {2024},
  url = {https://www.anthropic.com/research/model-context-protocol}
}

@inproceedings{wangevolving,
  title={EVOLVING ROLLOUTS: Harnessing Historical Experience for Web Agent Evolution in Reinforcement Learning},
  author={Wang, Sinuo and Wang, Piaohong and Qin, Tianrui and Song, Maojia and Chen, Qianben and Wang, Qiexiang and Zhou, Gengze and Zhang, Zeyu and Zhu, He and Shi, Dingfeng and others},
  booktitle={Forty-third International Conference on Machine Learning},
  year={2026}
}

@article{zhou2025livesearchbench,
  title={Livesearchbench: An automatically constructed benchmark for retrieval and reasoning over dynamic knowledge},
  author={Zhou, Heng and Yu, Ao and Fan, Yuchen and Shi, Jianing and Kang, Li and Geng, Hejia and Zhang, Yongting and Fan, Yutao and Wu, Yuhao and He, Tiancheng and others},
  journal={arXiv preprint arXiv:2511.01409},
  year={2025}
}

@misc{lu2026analystgroundtruthgrading,
  title={One Analyst Is Not Ground Truth: Grading Agent-Built Financial Models Against Observed Professional Practice},
  author={Jiacheng Lu and Yiming Li and Sinuo Wang and Wentao Zhao and Rui Sun and Cheng Hua and Tao Song and Hui Cai and Beidi Luan and Tianyi Ma and Qixuan Yang and Zhengze Wu and Jing Li and Daxin Jiang and Zuo Bai and Haibing Guan},
  year={2026},
  eprint={2607.24889},
  archivePrefix={arXiv},
  primaryClass={cs.LG},
  url={https://arxiv.org/abs/2607.24889},
}
\end{document}